# Scientific Machine Learning Benchmarks

Jeyan Thiyagalingam[*], Mallikarjun Shankar[†], Geoffrey Fox[‡], and Tony Hey[*]

## Abstract

The breakthrough in Deep Learning neural networks has transformed the use of AI and machine learning technologies for the analysis of very large experimental datasets. These datasets are typically generated by large-scale experimental facilities at national laboratories. In the context of science, scientific machine learning focuses on training machines to identify patterns, trends, and anomalies to extract meaningful scientific insights from such datasets. With a new generation of experimental facilities, the rate of data generation and the scale of data volumes will increasingly require the use of more automated data analysis.

At present, identifying the most appropriate machine learning algorithm for the analysis of any given scientific dataset is still a challenge for scientists. This is due to many different machine learning frameworks, computer architectures, and machine learning models. Historically, for modelling and simulation on HPC systems such problems have been addressed through benchmarking computer applications, algorithms, and architectures. Extending such a benchmarking approach and identifying metrics for the application of machine learning methods to scientific datasets is a new challenge for both scientists and computer scientists. In this paper, we describe our approach to the development of scientific machine learning benchmarks and review other approaches to benchmarking scientific machine learning.

## 1 Introduction

In the past decade, a sub-field of artificial intelligence (AI), namely Deep Learning (DL) neural networks (or deep neural networks, DNNs), has made significant breakthroughs in many scientifically and commercially important applications[1]. Such neural networks are themselves a subset of a wide range of machine learning (ML) methods (Figure 1.)

---

[*] Rutherford Appleton Laboratory, Science and Technology Facilities Council, Harwell Campus, United Kingdom. **t.jeyan@stfc.ac.uk; tony.hey@stfc.ac.uk**
[†] Oak Ridge National Laboratory, 1 Bethel Valley Road, Oak Ridge, TN 37831, USA. **shankarm@ornl.gov**
[‡] University of Virginia, Computer Science and Biocomplexity Institute, 994 Research Park Blvd, Charlottesville, Virginia, 22911, USA. **vxj6mb@virginia.edu**

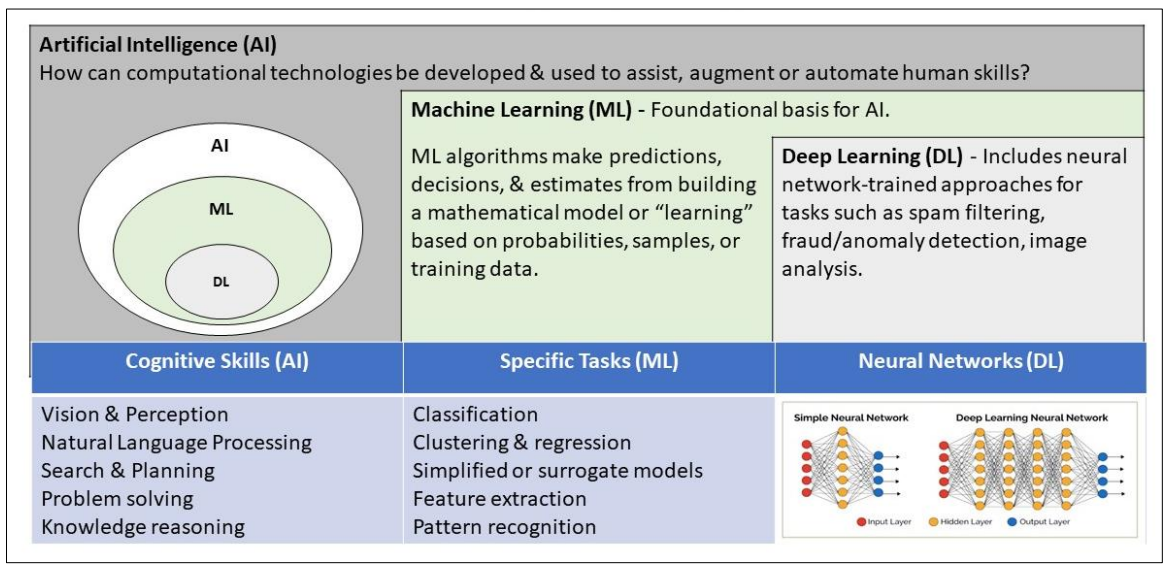

Figure 1: AI, Machine Learning and Deep Learning (Adopted from Workshop Report on Basic Research Needs for Scientific Machine Learning[2])

ML methods have been widely used for many years in several domains of science, but DNNs have been transformational and are gaining a lot of traction in many scientific communities[3]. Most of the national laboratories that host large-scale experimental facilities are now relying on DNN-based data analytic methods to extract scientific insights from their increasingly large datasets. A recent spectacular success is DeepMind's use of Deep Learning in their Alpha Fold-1 and Alpha Fold-2[4] solutions to the protein folding 'Grand Challenge'. This promises to transform much of biological science and open up exciting new research avenues. Other domains of science are exploring physical representations of the system with the data-driven learning ability of neural networks. Current developments are towards specialising these ML approaches to be more domain-specific and domain-aware[5–7], and aiming to connect the apparent 'black box' successes of DL networks with well-understood approaches from science.

The overarching scope of ML in science is very broad, including identifying patterns, anomalies, and trends from relevant scientific datasets, and using ML for classification and predicting of those patterns, clustering of data, and generating near-realistic synthetic data. There are three approaches for developing ML-based solutions, namely, supervised, unsupervised, and reinforcement learning. In supervised learning, the ML model is trained for a given task with examples. In order to have examples, the data used for training the ML model must contain the ground truth or labels. Supervised learning is therefore only possible when there is a labelled subset of the data. Once trained, the learned model can be deployed for real-time usage, such as pattern classification or estimation --- which is often referred to as *inference*. Because of the difficulty in generating labelled data for supervised learning, particularly for experimental datasets, it is often difficult to apply supervised learning directly. To circumvent this limitation, training is often performed on simulated data, which provides an opportunity to have relevant labels. However, the simulated data may not be representative of the real data and the model may therefore not perform satisfactorily when used for inferencing. The unsupervised learning technique, in contrast, does not rely on labels. A simple example of this technique is clustering, where the aim is to identify several groups of data points that have common features. Another example is identification of anomalies in data. Example algorithms include k-Means Clustering[8], Support Vector Machines (SVM)[9], or neural network-based autoencoders[10]. Finally, reinforcement learning relies on a trial-and-error approach to learn a given task with the learning system being positively rewarded whenever the system behaves correctly, and penalised whenever it behaved incorrectly[11]. Each of these learning paradigms have a large number of algorithms, and modern developmental approaches are often hybrid and use one of more of these techniques together. This leaves a very large choice of ML algorithms for any given problem.

In practice, the selection of an ML algorithm for a given scientific problem is more complex than just selecting one of the machine learning technologies and any particular algorithm. The selection of the most effective ML algorithm is based on many factors, including the type, quantity, and quality of the training data, the availability of labelled data, the type of problem being addressed (prediction, classification, and so on), the overall accuracy and performance required, and the hardware systems available for training and inferencing. With such a multi-dimensional problem consisting of a choice of ML algorithms, hardware architectures, and a range of scientific problems, selecting an optimal ML algorithm for a given task is not trivial. This constitutes a significant barrier for many scientists wishing to use modern ML methods in their scientific research.

In this paper we use suitable scientific ML benchmarks to develop guidelines and best practices to assist the scientific community in successfully exploiting these methods. Moreover, developing such guidelines and best practices at the community level will not only benefit the science community but also highlight where further research into ML algorithms, computer architectures, and software solutions for using ML in scientific applications is needed.

Such guidelines and best practices need to be based on real-world application examples and relevant data. For instance, demonstrating the success of a specific ML technique on a specific scientific problem will assist researchers in applying the technique to similar problems. We refer to the development of guidelines and best practices as *Benchmarking*. In our case, this is very specific to ML techniques applied to scientific datasets. The applications used to demonstrate the guideline and best practices are referred to as *Benchmarks*.

The notion of benchmarking computer systems and applications has been a fundamental cornerstone of computer science, particularly for compiler, architectural and system development, with a key focus on using benchmarks for ranking systems, such as the Top500 or Green500[12–16]. However, our notion of scientific ML benchmarking has a different focus. Firstly, these machine learning benchmarks can be considered as blueprints for use on a range of scientific problems, and hence are aimed at fostering the use of ML in science more generally. Secondly, by using these ML benchmarks, a number of aspects in an ML ecosystem can be compared and contrasted. For example, it is possible to rank different computer architectures for their performance, or to rank different ML algorithms for their effectiveness. Thirdly, these ML benchmarks are accompanied by relevant dataset(s) on which the training and/or inference will be based. This is different to conventional benchmarks for high-performance computing (HPC) where there is little dependency on datasets. The establishment of a set of open curated datasets with associated ML benchmarks is therefore an important step for scientists to be able to effectively utilise ML methods in their research and also to identify further directions for ML research.

In this paper, we first discuss what we mean by scientific machine learning benchmarks, the scope of such benchmarks, and the challenges in creating such benchmarks. We then review a number of benchmarking initiatives in light of this discussion. The paper is organised as follows. In Section 2, we discuss the primary considerations in designing benchmarks to advance the application of ML methods for scientific research along with relevant examples. We then define the scope and challenges around establishing such scientific machine learning benchmarks in Section 3. In Section 4, we review a number of ML benchmarking initiatives in light of our discussions in Sections 2 and 3. We then discuss SciMLBench, one of the most recent and versatile scientific ML benchmarking initiatives, in Section 5. We summarise our findings and conclusions in Section 6.

## 2. Machine Learning Benchmarks for Science

### 2.1 Elements of a Benchmark for Science

As discussed above, a scientific ML benchmark is underpinned by a scientific problem and should have two elements: (a) the dataset on which this benchmark is trained or inferenced upon, and (b) a reference implementation, which can be in any programming language (e.g., Python, C++). The scientific problem can be from many different scientific domains. A collection of such benchmarks can make up a benchmark suite as illustrated in Figure 2.

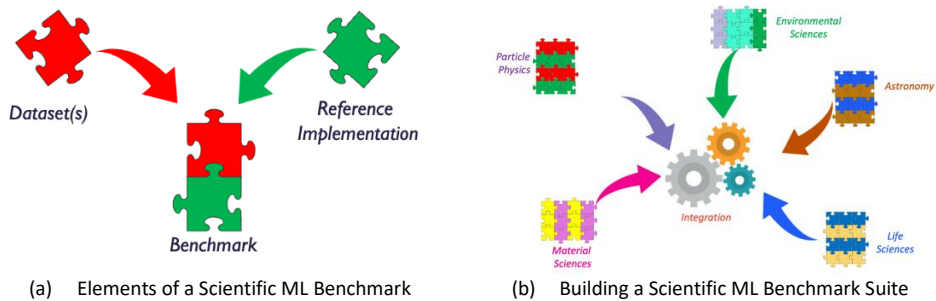

(a)   Elements of a Scientific ML Benchmark     (b)   Building a Scientific ML Benchmark Suite

Figure 2: Notion of an ML Benchmark and a Benchmark Suite

### 2.2 Focus of Benchmarking

There are three separate aspects of scientific benchmarking that apply in the context of ML benchmarks for science, namely scientific ML benchmarking, application benchmarking and system benchmarking. We illustrate these in Figure 3.

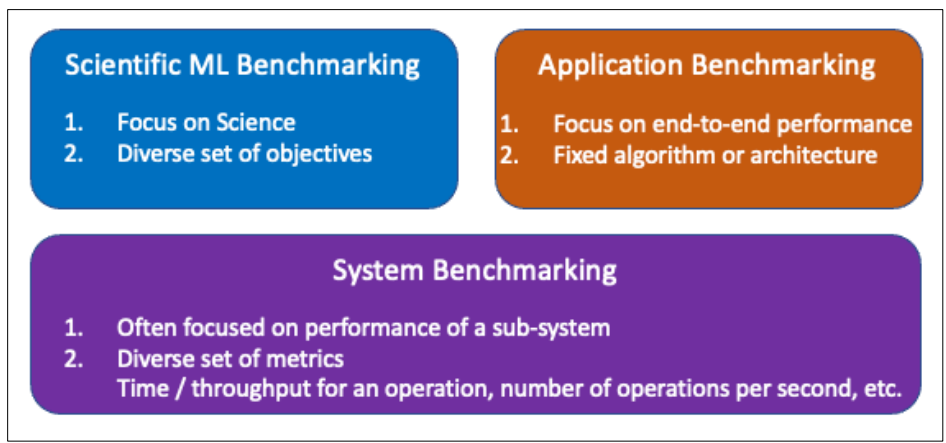

Figure 3: Different focus areas of benchmarking

- **Scientific ML Benchmarking:** This is concerned with algorithmic improvements that help reach the scientific targets specified for a given dataset. Here we wish to test algorithms and their performance on fixed data assets, typically with the same underlying hardware and software environment. This type of benchmark is characterized by the dataset together with some specific scientific objectives. The data is obtained from a scientific experiment and should be rich enough to allow different methods of analysis and exploration. Examples of metrics could include the F1 score for training accuracy and time-to-solution.
- **Application Benchmarking:** This aspect of ML benchmarks is concerned with exploring the performance of the complete ML application when using different hardware and software environments. A typical performance target would be time-to-solution. This type of benchmark is characterized by the ML application and data which can then be used to evaluate the performance of the overall system (hardware, software libraries, runtime environments, filesystems, etc.) in the context of the given application and data. Examples of metrics could include a throughput measure (e.g., images per second), time-to-solution of the application, and investigation of the scaling properties of the application.
- **System Benchmarking:** This is concerned with investigating performance effects of the system hardware architecture on improving the scientific outcomes/targets. These benchmarks have similarities with application benchmarks, but they are characterized by primarily focusing on a specific operation that exercises a particular part of the system, independent of the broader system environment. Suitable metrics could be time-to-solution, the number of floating-point operations per second (FLOP/s) achieved, or aspects of network and data movement performance.

**2.3 Examples of Scientific Machine Learning Benchmarks**

Scientific ML Benchmarks are ML applications that solve a particular scientific problem from a specific scientific domain. For example, this can be as simple as an application that classifies the experimental data in some way, or as complex as inferring the properties of a material from neutron scattering data. Some examples are:

1. Inferring the structure of multi-phase materials from X-ray diffuse multiple scattering data. Here, the machine learning is used to automatically identify the phases of materials using classification. This is an example from the materials science domain.
2. Estimating the photometric redshifts of galaxies from survey data. Here, the machine learning is used for estimation. This example is drawn from astronomy[17].
3. Clustering of microcracks in a material using X-ray scattering data. Here, the machine learning uses an unsupervised learning technique.
4. Removing noise from microscope data to improve the quality of images. Machine learning is used for its capability to perform high quality regression of pixel values.

We provide more detailed examples in later sections.

**3 Benchmarking Process**

Although it is possible to provide a collection of ML-specific scientific applications (with relevant datasets) as benchmarks for any of the purposes mentioned above, the exact process of benchmarking requires the following processes:

- **Metrics of Choice**: First, depending on the focus (see Section 2), the exact metric by which different benchmarks are compared may vary. For example, if science is the focus, then this metric may vary from benchmark to benchmark. However, if the focus is system-level benchmarking, it is possible to agree on a common set of metrics that can span across a range of applications.
- **Framework**: Providing just a collection of disparate applications without a coherent mechanism for evaluation will require users perform several fairly complex benchmarking operations for their specific goals. Ideally, therefore, the benchmark suite should also offer a framework that helps in achieving these goals and that unifies aspects that are common to all applications in the suite, such as portability, flexibility, and logging.
- **Reporting and Compliance**: Finally, how these results are reported is important. In many cases, a benchmark framework as discussed above addresses this concern. However, there are often some specific compliance aspects that must be followed to ensure that the benchmarking process is carried out fairly across different hardware platforms.

There are also a number of challenges which need to be addressed when dealing with the development of ML benchmarks. These are:

- **Data**: In the previous section, we highlighted the significance of data when using ML for scientific problems. The availability of curated, large-scale, scientific datasets - which can be either experimental or simulated data – is the key to developing useful ML benchmarks for science. Although much scientific data is openly available, the curation, maintenance, and distribution of large-scale datasets for public consumption is a challenging process.  A good benchmarking suite needs to provide a wide range of curated scientific datasets coupled with the applications. Reliance on external datasets has the danger of not having full control or even access to those datasets.
- **Distribution**: A scientific ML benchmark constitutes a reference implementation and relevant dataset and both these must be available to the users. Since realistic dataset sizes can be in the terabytes (TB) range, the access and downloading of the datasets is not always straightforward.
- **Coverage**: Benchmarking is a very broad topic and providing benchmarks to cover the different focus areas highlighted above, across a range of scientific disciplines, is not a trivial task. A good benchmark suite should provide a good coverage of methods and goals and should be extensible.
- **Extensibility:** Although the notion of scientific machine learning benchmarks can be interesting for scientists, it can be very time-consuming to develop benchmark-specific codes. If the original scientific application needs substantial refactoring to be converted into a benchmark, this will not be an attractive option for scientists. Any benchmarking framework should therefore try to minimise the amount of code refactoring required for conversion into a benchmark.

## 4. Review of Benchmarking Initiatives

Comparing different ML techniques is not a new requirement and is increasingly becoming common in ML research. In fact, this approach has been fundamental for the development of various ML techniques. For example, the ImageNet[18,19] dataset spurred a competition to improve computer image analysis and understanding and been widely recognized for driving innovation in deep learning. However, providing a blueprint of applications, guidelines, and best practices in the context of scientific machine learning is a relatively new requirement. There have been a number of efforts on this aspect that address some of the challenges we highlighted in Section 3.  In this brief review of these benchmarking initiatives, we explicitly exclude conventional benchmarking activities in other areas of computer science, such as benchmarks for HPC systems, compilers, and sub-systems such as memory, storage, and networking[12,20].

Instead of giving an exhaustive technical review covering very fine-grained aspects, we give a very high-level review of the various ML benchmark initiatives here, focussing on the requirements discussed in Sections 2 and 3. We shall therefore cover the following aspects in our review:

1. Benchmark Focus: Science, Application (End-to-End), and System.
2. Benchmark Process: Metrics, Framework, and Reporting & Compliance.
3. Benchmark Challenges: Data, Distribution, Coverage, and Extensibility.

In the context of ML benchmarking, there are a several initiatives such as Deep500[21], RLBench[22], CORAL-2[23], DAWNBench[24], AI Bench[25], MLCommons[26], SciML-Bench[27], as well as widely recognised Community Competitions (such as those organized by Kaggle[28]). We review these initiatives below and note that a specific benchmarking initiative may or may not support all the aspects listed above or, in some cases, may only offer partial support.

## 4.1 Deep 500

The Deep500[21] initiative proposes a customizable and modular software infrastructure to aid in comparing the wide range of deep learning frameworks, algorithms, libraries, and techniques. The key idea behind Deep500 is its modular design, where deep learning is factorized into four distinct levels: operators, network processing, training, and distributed training. While this approach aims to be neutral and overarching, and able to accommodate a wide variety of techniques and methods, the process of mapping a code to a new framework has impeded its adoption for new benchmark development. Furthermore, despite its key focus on deep learning, neural networks, and a very customisable framework, benchmarks or applications are not included by default and are left for the end user to provide, as is support for reporting. The main limitation is the lack of a suite of representative benchmarks.

## 4.2 RLBench

RLBench[22] is a benchmark and learning environment featuring hundreds of unique, hand-crafted tasks. The focus is on a set of tasks to evaluate new algorithmic developments around reinforcement learning, imitation learning, multi-task learning, geometric computer vision, and in particular, few-shot learning.

The tasks are very specific and can be considered as building blocks of large-scale applications. However, the environment currently lacks support for the classes of benchmarking discussed in section 2

## 4.3 CORAL-2

The CORAL-2[23] benchmarks are computational problems relevant to a scientific domain or to data science, and are typically backed by a community code. Vendors are then expected to evaluate and optimize these codes to demonstrate the value of their proposed hardware in accelerating computational science. This allows a vendor to rigorously demonstrate the performance capabilities and characteristics of a proposed machine on a benchmark suite that should be relevant for computational scientists. The machine learning and data science tools in CORAL-2 include a number of ML techniques across two suites, namely, the big data analytics (BDS) and deep learning (DLS) suites. While the BDS suite covers conventional ML techniques, such as principal components analysis (PCA), k-means clustering, and support vector machines (SVM), the DLS relies on the ImageNet[18,19] and CANDLE[29] benchmarks which are primarily used for testing scalability aspects rather than purely focussing on the science. Similarly, the BDS suite aims to exercise the memory constraints (PCA), computing capabilities (SVM), and/or both these aspects (k-Means) and is also concerned with communication characteristics. Although these benchmarks are oriented at machine learning, the constraints and benchmark targets are narrowly specified and emphasize scalability capabilities. The overall coverage of science in the CORAL-2 benchmark suite is quite broad but the footprint of the ML techniques is limited to the BDS and DLS suites and there is little focus on scientific data distribution for algorithm improvement.

## 4.4 AI Bench

The AI Bench initiative is supported by the International Open Benchmark Council (Bench Council)[25]. The Council is a non-profit international organization that aims to promote standardizing, benchmarking, evaluating, and incubating Big Data, AI, and other emerging technologies. The scope of AI Bench is very comprehensive and includes a broad range of internet services, including search engines, social networks, and e-commerce. The underlying ML-specific tasks in these areas include image classification, image generation, translation (image-to-text, image-to-image, text-to-image, text-to text), object detection, text summarisation, advertising, and natural language processing. The relevant datasets are open, and the primary metric is system performance for a fixed target. AI Bench currently lacks a specific science focus and a framework but the AI Bench environment does enforce some level of compliance for reporting ranking information of hardware systems.

**4.5 DawnBench**

DawnBench[24] is a benchmark suite for end-to-end deep learning training and inference. The end-to-end aspect here is ideal for application and system level benchmarking. Instead of focussing on model accuracy, DawnBench provides common deep learning workloads for quantifying training time, training cost, inference latency, and inference cost across different optimization strategies, model architectures, software frameworks, clouds, and hardware. There are two key benchmarks in the suite – image classification (using the ImageNet and CIFAR10 [30]datasets) and Natural Language Processing-based Question Answering (based on the Stanford Question Answering Dataset or SQuAD) that covers both training and inference. DawnBench does not offer a notion of a framework and does not have a focus on science. With key metrics around time and cost (for training and inference), DawnBench is predominantly targeted towards end-to-end system and algorithmic performance. Although the datasets are public and open, no distribution mechanisms have been adopted by DawnBench.

**4.6 Benchmarks from MLCommons Working Groups**

MLCommons is an international initiative aimed at improving all aspects of the ML landscape and covers benchmarking, datasets, and best practices. The consortium has several working groups around various different focii for ML applications. Among these working groups, two are of interest here: HPC and Science. The MLCommons HPC benchmark[26] suite focuses on scientific applications that use ML, and especially on deep learning (DL) at HPC scale. The codes and data are specified in such a way that execution of the benchmarks on supercomputers will help understand detailed aspects of system performance. The focus is on performance characteristics particularly relevant to HPC applications such as model-system interactions, optimization of the workload execution, and reducing throughput or execution bottlenecks. The HPC orientation also drives this effort towards exploration of benchmark scalability.

By contrast, the MLCommons Science benchmark[31] suite focuses specifically on the application of ML methods to scientific applications and includes application examples across several scientific domains. However, the suite currently lacks a supportive framework for running the benchmarks but, as with the rest of the MLCommons, does enforce compliance for reporting of the results. The benchmarks cover the three areas of benchmarking - science, application, and system – as described in Section 2.

### 4.7 SciMLBench

The Scientific Machine Learning Benchmark suite - or SciMLBench[27] – has been developed by the Scientific Machine Learning Research Group at the Rutherford Appleton Laboratory in the UK. The suite of benchmarks is specifically focussed on scientific machine learning and covers nearly every aspect of the cases we discussed in Sections 2 and 3. We provide a fuller description of the SciMLBench in Section 5.

### 4.8 Community Competitions

Although community-based competitions, such as those organised by Kaggle[28], can be seen as a benchmarking activity, these competitions are do not have a coherent methodology or a controlled approach for developing benchmarks. In particular the competitions do not provide a framework for running the benchmarks nor do they consider data distribution methods. Each competition is tailored on its own and relies on its own dataset, set of rules, and compliance metrics. The competitions are also typically short-lived. Although such challenge competitions can provide a blueprint for using ML technologies for specific research communities, they are unlikely to deliver best practices or guidelines for the long-term.

## 5. The SciML-Bench Approach

The SciMLBench approach was developed by the Scientific Machine Learning Group at the Rutherford Appleton Laboratory in collaboration with researchers at Oak Ridge National Laboratory and at the University of Virginia. Among all the approaches we reviewed in Section 4, only the SciMLBench benchmark suite addresses nearly all of the concerns raised in Section 2. To the best of our knowledge, the SciMLBench approach is unique in its versatility compared to the other approaches and its key focus is on scientific machine learning.

### 5.1 Core Components
The SciMLBench has three components, namely:

- **Benchmarks**: The benchmarks are machine learning applications performing a specific scientific task, written in Python. These are included by default, and users are not required to find applications on their own. On the scale of micro-apps, mini-apps, and apps, these codes are full-fledged applications. Each benchmark aims to solve a specific scientific problem (see Section 2.3 for science examples).
- **Datasets**: Each benchmark relies on one or more datasets which can be used, for example, for training and/or inferencing. These datasets are open, task- or domain-specific, and FAIR compliant. Since most of these datasets are large, they are hosted separately, on one of our Lab servers (or mirrors), and are automatically or explicitly downloaded on demand.
- **Framework**: The framework serves two purposes. Firstly, at the user level, it facilitates an easier approach to the actual benchmarking, logging, and reporting of the results. Secondly, at the developer level, it provides a coherent API for unifying and simplifying the development of ML benchmarks.

The SciML framework is the basic fabric upon which the benchmarks are built. It is both extensible and customizable and offers a set of application programming interfaces (APIs). These APIs enable easier development of benchmarks based on this framework. The framework is architecture-independent and the minimum system requirement is determined by the specific benchmark. There is a built-in logging mechanism that captures all potential system-level and benchmark-level outputs during execution. The central component that links benchmarks, datasets, and the framework is a configuration tool that the framework relies on. The most attractive part of the framework is the possibility of simply using existing codes as benchmarks with only a few API calls necessary to register the benchmarks. Finally, the framework is designed with scalability in mind, so that benchmarks can be run on any computer ranging from a single system to a large-scale supercomputer.

## 5.2 Benchmarks and Datasets

The currently released version of SciMLBench has three benchmarks with their associated datasets. The benchmarks from this release represent scientific problems drawn from material sciences and from environmental sciences, namely:

1. Diffuse Multiple Scattering (DMS_Structure, Material Sciences): This benchmark uses machine learning for classifying the structure of multi-phase materials from X-ray scattering patterns. More specifically, the machine learning based approach enables automatic identification of phases. This application is particularly useful for the material science community as diffuse multiple scattering allows investigation of multi-phase materials from a single measurement – something not possible with standard X-ray experiments. However, manual analysis of the data can be extremely laborious, involving searching of patterns to identify important motifs (triple intersections) that allow for inference of information. This is a multi-label classification problem (as opposed to a binary classification problem as in the Cloud masking example discussed below). The benchmark relies on a simulated dataset of size 8.6GB with three-channel images of resolution 487x195 pixels.

2. Cloud Masking (SLSTR_Cloud, Environmental Sciences): Given a set of satellite images, the challenge for this benchmark to classify each pixel of each satellite image as either cloud or as non-cloud (clear sky). This problem is known as 'cloud masking' and is crucial for several important applications in earth observation. In a conventional, non-ML setting, this task is typically performed using either thresholding or Bayesian methods. The benchmark exercises deep learning and includes two datasets, DS1-Cloud and DS2-Cloud, with sizes of 180GB and 1.2TB, respectively. The datasets contain multi-spectral images with resolution of 2400 x 3000 pixels or 1200 x 1500 pixels.

3. Electron Microscopy Image Denoising (EM_Denoise): This benchmark uses machine learning for removing noise from electron microscopic images. This improves the signal to noise ratio of the image and is often used as a precursor to more complex techniques such as surface reconstruction or tomographic projections. Effective denoising can facilitate low-dose experiments in producing images with a quality comparable that obtained in high-dose experiments. Likewise, greater time resolution can also be achieved with the aid of effective image denoising procedures. This benchmark exercises complex deep learning techniques on a simulated dataset of size 5GB, consisting of 256x256 images covering noised and denoised (ground truth) datasets.

The next release of the suite will include several more examples from various other domains with large datasets.

### 5.3 Benchmark Focus

With the full-fledged capability of the framework to log all activities, and with a detailed set of metrics, it is possible for the framework to collect a wide range of performance details that can later be used for deciding the focus. For example, SciMLBench can be used for science benchmarking (to improve scientific results through different ML approaches), application-level benchmarking, and system-level benchmarking (gathering end-to-end performance including IO and network performance). This is made possible thanks to the framework's detailed logging mechanisms.

### 5.4 Benchmarking Process

With the framework handling most of the complexity of collecting performance data, there is an excellent opportunity to cover a wide range of metrics (even retrospectively after the benchmarks have been run) and have the ability to control the reporting and compliance through controlled runs. However, it is worth noting that although the framework supports and collects a wide range of runtime and science performance aspects, the choice is left to the user to decide the ultimate metrics to be reported. For example, the performance data collected by the framework can be used to generate a final figure of merit to compare different ML models or hardware systems for the same problem.

### 5.5 Data Curation and Distribution

SciMLBench employs a carefully designed curation and distribution mechanism:

a) Each benchmark has one or more associated datasets. These benchmark-dataset associations are specified through a configuration tool which is not only framework friendly but also interpretable by scientists.

b) As the scientific datasets are usually large, they are not maintained along with the code. Instead, they are maintained in a separate object storage, whose exact locations are visible to the benchmarking framework and to users.

c) Users downloading benchmarks will only download the reference implementations (code) and not the data. This enables fast downloading of the benchmarks and the framework. Since not all datasets will be of interest to everyone, this approach prevents unnecessary downloading of large datasets.

d) The framework takes the responsibility for downloading datasets on demand or when the user launches the benchmarking process.

This process is illustrated Figure 4 below.

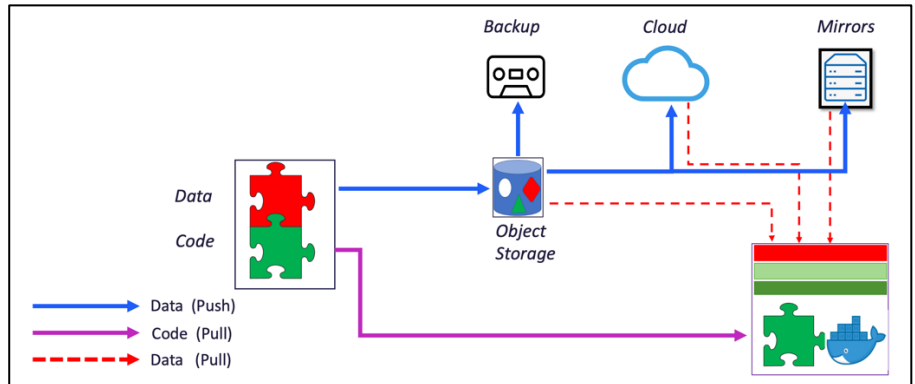

Figure 4: Moving the benchmark datasets to the evaluation point

In addition to these basic operational aspects, the benchmark datasets are stored in an object storage to enable better resiliency and repair mechanisms compared to simple file storage. The datasets are also mirrored in several locations to enable the framework to choose the data source closest to the location of the user. The datasets are also regularly backed-up as they constitute valuable digital assets.

## 5.6 Extensibility and Coverage

The overall design of the SciMLBench supports several user scenarios: the ability to add new benchmarks with little knowledge of the framework, ease-of-use, platform interoperability, and ease of customization. The overall design relies on two application programming Interface (API) calls which are illustrated with a number of toy examples as well as with some practical examples.

## 6 Outlook and Conclusions

In this paper, we have highlighted the need for scientific machine learning benchmarks and explained how they differ from conventional benchmarking initiatives. Furthermore, we have highlighted the challenges in developing a suite of useful scientific machine learning benchmarks. These challenges span a number of issues ranging from the intended focus of the benchmarks (Section 2.2) and the benchmarking processes (section 2.3), to challenges around actually developing a useful ML benchmark suite (section 3). A useful scientific machine learning suite must therefore go beyond just providing a disparate collection of ML-based scientific applications. The critical aspect here is to provide support for end users not only to be able to effectively use the ML benchmarks but also to enable them to develop new benchmarks and extend the suite for their own purposes.

Our review covered a number of contemporary efforts for developing ML benchmarks. Only a subset of these efforts has a focus of machine learning for scientific applications. Furthermore, we also noted that nearly all of these initiatives do not consider the problem of the efficient distribution of large datasets. The majority of the approaches rely on externally sourced datasets with the implicit assumption that users will take care of the data issues. We concluded the paper with a more detailed review of our SciMLBench initiative which includes a benchmark framework that not only addresses the majority of these concerns but is also designed for easy extensibility.

The characteristics of these ML benchmark initiatives are summarised in Table 1 below. In qualitatively assessing how far each approach addresses the concerns discussed in Section 2, we have indicated whether they offer no support (none), or partial or questionable support (partial) or fully support the concern (full). To ease readability and understanding, we use a traffic-light coloured glyphs of red, amber and green (●, ● and ●) for none, partial or full support, respectively.

Table 1: Overall Assessment of Various Scientific Machine Learning Benchmarking Approaches

| Relevant Work | Focus | | | Process | | | Challenges | | | |
|---|---|---|---|---|---|---|---|---|---|---|
| | Scientific | Application | System | Metrics | Framework | Reporting | Data | Distribution | Coverage | Extensibility |
| Deep 500 | ● | ● | ● | ● | ● | ● | ● | ● | ● | ● |
| RL-Bench | ● | ● | ● | ● | ● | ● | ● | ● | ● | ● |
| CORAL-2 (DLS/BDS) | ● | ● | ● | ● | ● | ● | ● | ● | ● | ● |
| AI-Bench | ● | ● | ● | ● | ● | ● | ● | ● | ● | ● |
| DawnBench | ● | ● | ● | ● | ● | ● | ● | ● | ● | ● |
| MLCommons-Science | ● | ● | ● | ● | ● | ● | ● | ● | ● | ● |
| SciML-Bench | ● | ● | ● | ● | ● | ● | ● | ● | ● | ● |
| Community Competitions | ● | ● | ● | ● | ● | ● | ● | ● | ● | ● |

The table shows that the benchmarking community has several issues to address to ensure that the scientific community is equipped with right set of tools to become more efficient in leveraging the use of ML technologies in science.

**Contributions:** JT, MS, GF and TH conceptualised the idea of Scientific Benchmarking. JT designed the SciMLBench framework, data architecture and conceptualised the overarching set of features. TH has overseen the overall developmental efforts along with JT, MS and GF. All have contributed towards the writing of the manuscript.

**Acknowledgements:** We would like to thank Samuel Jackson, Kuangdai Leng, Keith Butler and Juri Papay from the Scientific Machine Learning Research Group at the Rutherford Appleton Laboratory, Junqi Yin and Aristeidis Tsaris from Oak Ridge National Laboratories, and the MLCommons Science Working Group for valuable discussions. This work was supported by Wave 1 of The UKRI Strategic Priorities Fund under the EPSRC Grant EP/T001569/1, particularly the "AI for Science" theme within that grant, by the Alan Turing Institute, and by the Benchmarking for AI for Science at Exascale (BASE) project under the EPSRC Grant EP/V001310/1.